%% file: root.tex
\documentclass[letterpaper, 10 pt, conference]{ieeeconf}  %
\usepackage{xcolor}

\IEEEoverridecommandlockouts                              %

\usepackage{graphics} %
\usepackage{epsfig} %
\usepackage{mathptmx} %
\usepackage{times} %
\usepackage{amsmath} %
\usepackage{amssymb}  %

\usepackage{booktabs}

\usepackage{float}    %
\usepackage{graphicx} %

\usepackage{multirow}
\usepackage{pifont}

\usepackage{amsmath,amssymb,amsfonts}
\usepackage{algorithm}
\usepackage[noend]{algpseudocode}
\algrenewtext{For}[1]{\textbf{For} #1}
\algrenewtext{EndFor}{\textbf{end for}}
\usepackage{amsmath}
\usepackage{algorithm}
\usepackage{algpseudocode}

\title{\LARGE \bf
HiD$^{2}$: A Trajectory Generator for \underline{Hi}gh-\underline{D}ensity Traffic and \underline{D}iverse Agent-Interaction Scenarios
}

\author{Ruining Yang, Yi Xu, Yixiao Chen, Yun Fu, and Lili Su%
\thanks{Department of Electrical and Computer Engineering,
        Northeastern University, Boston, MA 02115, USA}%
}

\begin{document}

\maketitle
\thispagestyle{empty}
\pagestyle{empty}

\input{sec/0_abstract}
\input{sec/1_intro}

\input{sec/2_related_work}

\input{sec/3_method}

\input{sec/4_experiments}

\input{sec/5_conclusion}

\vfill\pagebreak
\bibliographystyle{IEEEtran}
\bibliography{references}

\addtolength{\textheight}{-12cm}   %

\end{document}

%% file: sec/0_abstract.tex
\begin{abstract}
Accurate trajectory prediction is fundamental to autonomous driving, as it underpins safe motion planning and collision avoidance in complex environments. 
However, existing benchmark datasets suffer from a pronounced long-tail distribution problem, 
with most samples drawn from low-density scenarios and simple straight-driving behaviors. 
This underrepresentation of high-density scenarios and safety critical maneuvers such as lane changes, overtaking and turning is an obstacle to model generalization and leads to overly optimistic evaluations.
To address these challenges, we propose a novel trajectory generation framework that simultaneously enhances scenarios density and enriches behavioral diversity. 
Specifically, our approach converts continuous road environments into a structured grid representation that supports fine-grained path planning, explicit conflict detection, and multi-agent coordination. 
Built upon this representation, we introduce behavior-aware generation mechanisms that combine rule-based decision triggers with Frenet-based trajectory smoothing and dynamic feasibility constraints. 
This design allows us to synthesize realistic high-density scenarios and rare behaviors with complex interactions that are often missing in real data. 
Extensive experiments on the large-scale Argoverse 1 and Argoverse 2 datasets demonstrate that our method significantly improves both agent density and behavior diversity, while preserving motion realism and scenario-level safety. Our synthetic data also benefits downstream trajectory prediction models and enhances performance in challenging high-density scenarios.

\end{abstract}

%% file: sec/1_intro.tex
\section{Introduction}

\begin{figure}[t]
    \centering
    \includegraphics[width=0.48\textwidth]{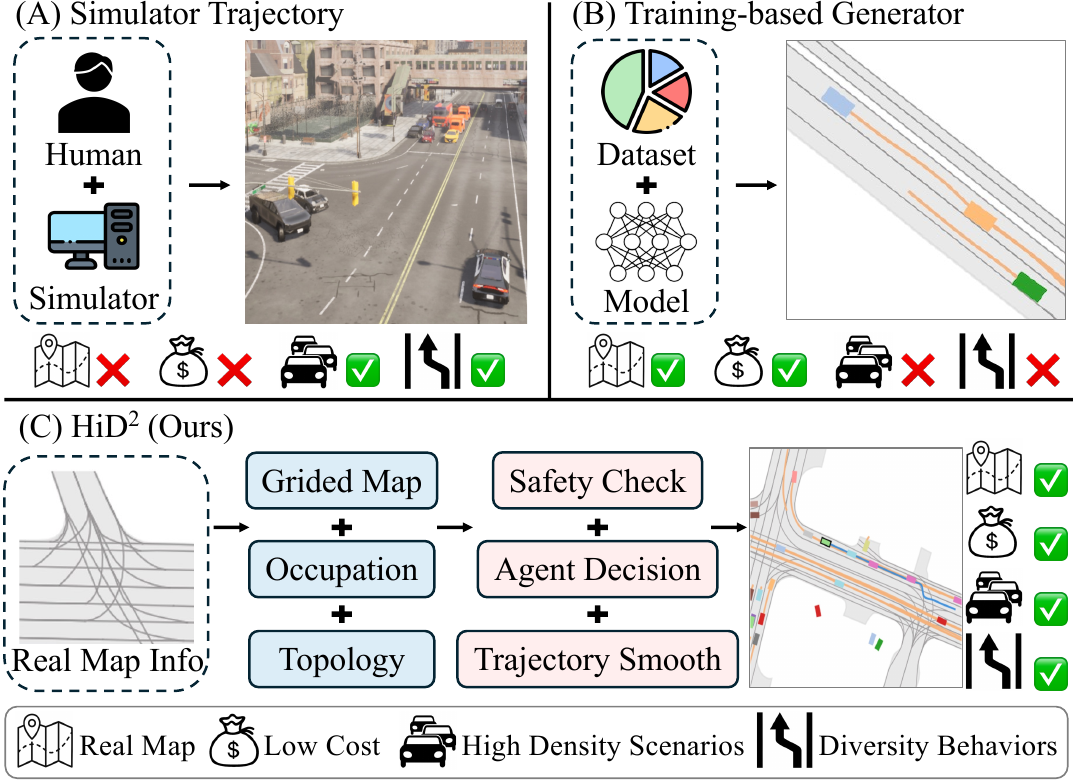}
    \caption{Comparison of different trajectory generation methods. \textbf{(A) Simulator generation}: Trajectories are generated through manual operation in a simulator, which can synthesize diverse behaviors, but lacks the constraints of a real map. \textbf{(B) Training-based methods}: Rely on raw datasets and models, making it difficult to generate rare behaviors in the tail. \textbf{(C) Our HiD$^{2}$ method}: Leveraging real maps and agent information, it generates high-density scenarios and diverse rare behaviors, effectively alleviating the long-tail distribution problem without requiring extensive manual effort.} 
    \label{fig:fig1}
    \vskip -1.2ex 
\end{figure}

Accurate trajectory prediction is fundamental to ensuring that autonomous vehicle can operate smoothly and safely in complex and uncertain environments~\cite{wang2025risk,yang2024towards}.  
Recent advances in artificial intelligence, both in model architectures such as transformers~\cite{vaswani2017attention,ngiam2021scene} and in hardware acceleration (e.g. modern GPUs), have substantially improved the speed and accuracy of prediction models. 
However, unlike conventional machine learning tasks, trajectory prediction faces unique challenges: the data are inherently multimodal and costly to annotate~\cite{wilson2023argoverse,ettinger2021large}, making dataset curation difficult and expensive.
This data bottleneck is becoming an obstacle to progress.
In particular, many benchmark datasets~\cite{chang2019argoverse,huang2018apolloscape,caesar2020nuscenes} suffer from long-tailed distributions in two key aspects: 
\begin{itemize}
    \item \textbf{Agent density imbalance:} 
    The number of traffic participants in a scenario can range from just a few to over 80~\cite{yang2024sstp}. Yet the distribution is skewed, most scenarios involve low to moderate density, while high-density scenarios are severely underrepresented.

    \item \textbf{Limited interaction diversity:} 
    The majority of trajectories correspond to straight driving, where agent interactions are minimal. In contrast, safety-critical maneuvers such as lane changes, overtaking, and sharp turns, where complex interactions arise, are relatively scarce.

\end{itemize}

These long-tailed issues hinder the generalization of the prediction models in critical scenarios where accurate behavior is most needed.
They also lead to overly optimistic evaluations that models trained and tested on low-density or simple-interaction scenarios may appear competent but fail in rare yet safety-critical situations.
For example, when a vehicle suddenly makes an unexpected right lane change with little clearance, a prediction model must correctly anticipate the maneuver to avoid collision.
Moreover, such challenges are particularly relevant for advanced perception models with attention mechanisms~\cite{zhou2022hivt,tang2024hpnet,wang2024towards}.
High agent density increases the computational complexity of attention, while diverse interactions directly affect attention allocation and the ability to capture dependencies.
As a result, benchmarks dominated by easy cases may overstate a model’s reliability, obscuring its weaknesses in rare but high-risk scenarios.

In this paper, inspired by the success of data generation in computer vision~\cite{shin2023fill,bansal2023leaving}, we tackle the long-tailed challenges in trajectory prediction through data generation.
Prior approaches on trajectory data generation can be broadly grouped into two categories: simulation-based and learning-based.
Simulation-based methods~\cite{dosovitskiy2017carla,yang2025trajectory} offer flexibility in generating arbitrary trajectories under physical constraints.
However, they require extensive manual effort for operation and post-filtering (e.g., removing collisions and off-road cases), and the resulting traffic scenarios often lack realism with respect to road geometry and traffic dynamics. 
Learning-based methods~\cite{feng2022trafficgen,tan2023language,xia2024language} typically rely on interpolation or perturbation of existing trajectories. 
While they can partially mitigate density imbalance by adding vehicles or extending fragmented tracks~\cite{feng2022trafficgen}, the resulting interactions remain limited.
This is not surprising, because from an information-theory perspective, most learning-based methods recycle existing patterns without introducing fundamentally new information.

To overcome these limitations, we introduce HiD$^{2}$, a trajectory generation framework designed for generating \underline{Hi}gh-\underline{D}ensity and \underline{D}iverse scenarios. 
HiD$^{2}$ synthesizes scenarios within real-world maps and produces diverse driving behaviors such as lane changes, overtaking, and sharp turns behaviors that are underrepresented yet safety-critical.
Our contributions are summarized as follows:
\begin{itemize}
    \item We propose HiD$^{2}$, which converts continuous road environments into a structured grid representation, enabling fine-grained trajectory generation that adheres to traffic rules and avoids collisions.
    On top of this, we design dedicated behavior-generation mechanisms that integrate rule-based triggers, Frenet-based smoothing, and dynamic feasibility constraints, ensuring that trajectories are both geometrically smooth and physically realistic.

    \item Through comprehensive evaluations, we demonstrate that HiD$^{2}$ preserves agent-level realism and scenario-level safety, while systematically enriching the diversity of high-density and rare-behavior cases, thereby filling critical gaps in existing datasets.  
  
    \item We show that augmenting training sets with HiD$^{2}$ data consistently improves the robustness of state-of-the-art trajectory prediction models, especially in high-density and interaction-heavy scenarios, compared to training on the original datasets alone. 
\end{itemize}

%% file: sec/2_related_work.tex
\section{Related Work}

\subsection{Long-tailed Distributions in Trajectory Prediction}
Trajectory prediction aims to infer an agent’s future motion based on its historical observations. Recent research has increasingly emphasized modeling complex multi-agent interactions, leading to significant progress in prediction methods~\cite{wang2024towards,giuliari2021transformer, xu2020spatial,zhou2022hivt,zhou2023query,tang2024hpnet,zhang2025decoupling,liang2020garden}.
Despite these advances, existing models still struggle in challenging scenarios~\cite{makansi2021exposing,pourkeshavarz2023learn,guo2024building}. One fundamental reason is that trajectory prediction datasets exhibit various forms of long-tailed distributions, such as imbalanced agent counts per scenario and a lack of behavioral diversity. To mitigate the impact of these imbalances, several strategies have been proposed.
FEND~\cite{wang2023fend} introduces a distribution-aware contrastive objective to align tail samples with dominant ones in the embedding space. HiVT-Long~\cite{zhou2022long}  incorporates uncertainty modeling to explicitly handle prediction errors in long-tail scenarios, while SSTP~\cite{yang2024sstp} uses gradient-based influence extraction and submodular selection to increase the representation of tail data.
While these approaches have shown promise, they remain confined to reusing. When critical tail cases are almost absent from the dataset, no amount of reusing existing data or perturbation can compensate for what does not exist.
In contrast, our work tackles the root cause by directly generating new scenarios on real HD maps, thereby addressing both key aspects of the long-tail problem: high agent density and behavioral diversity.

\subsection{Trajectory Generation}
Trajectory generation plays a crucial role in addressing the long-tailed problem in trajectory prediction.
Traditional simulator-based approaches~\cite{maroto2006real,casas2010traffic,papaleondiou2009trafficmodeler,erdmann2015sumo,fellendorf2010microscopic,lopez2018microscopic} rely on manual operations in virtual environments to generate trajectories. They can synthesize basic behaviors such as lane keeping, following, or simple interactions, but they are time-consuming and labor-intensive~\cite{yang2025trajectory}, and lack the ability to model real-world maps and traffic flow constraints. 

To generate higher-fidelity trajectories based on real maps, learning-based approaches~\cite{suo2021trafficsim,tan2021scenegen,xu2022bits,sun2022intersim,zhong2022guided,zhong2023language} have rapidly developed, leveraging real-world trajectory data to train generative models. TrafficGen~\cite{feng2022trafficgen} can synthesize a specified number of vehicles and their trajectories on a blank map. CTG~\cite{zhong2022guided} and CTG++~\cite{zhong2023language} introduce diffusion models and large language models to generate trajectory data. LCTGen~\cite{tan2023language} directly generates traffic scenarios through natural language descriptions without relying on historical data. InteractTraj~\cite{xia2024language} explicitly models the relationships between intelligent agents to capture cooperative and competitive behaviors. While these methods have made progress, they rely on raw data and thus can only produce samples that closely mirror the original distribution, where tail cases are already scarce. 
In contrast, our approach overcomes this limitation by leveraging an interaction-aware grid representation and explicit behavior controllers to generate rare yet realistic and safety-critical trajectories.

%% file: sec/3_method.tex
\section{Method}
\begin{figure*}[t]
    \centering
    \includegraphics[width=\textwidth]{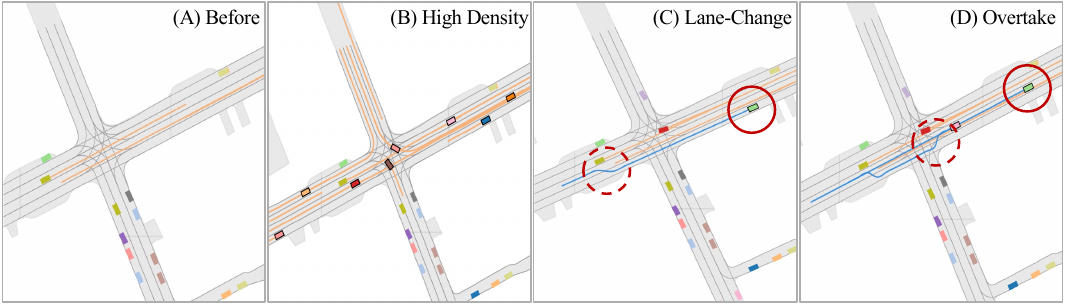}
    \caption{Visualization of generated trajectories under progressively complex driving scenarios. (A) Original trajectories in the baseline scenario. (B) Scenario with increased vehicle density, where the framework maintains robust trajectory generation despite tighter spacing and higher interaction frequency among vehicles. (C) Trajectory generation with lane-changing behavior, demonstrating the ability to adapt to dynamic intentions, negotiate surrounding traffic, and ensure collision-free maneuvering. (D) Trajectory generation with lane-changing and overtaking behaviors, highlighting the ability to implement competitive driving strategies and generate realistic multi-agent interactions in complex environments.}
    \label{fig:vis}
\end{figure*}

In this section, we formally describe how we generate traffic scenes with high agent density and complex agent-agent interactions on real-world maps. 

\subsection{Preliminary}
We model the road environment as a combination of a static high-definition (HD) map $\mathcal{M}$ and a set of dynamic agents $\mathcal{A}$. 
The static map $\mathcal{M}$ provides lane-level topology and geometry. 
Formally, it consists of a set of lanes $\mathcal{L} = \{l_1, l_2, \dots, l_{|\mathcal{L}|}\}$. 
Each lane $l$ is represented by an ordered sequence of centerline points $\mathbf{c}_l = \{c_0, c_1, \dots, c_{n_l}\}$, where $c_i \in \mathbb{R}^2$ is a 2D Cartesian coordinate. 
The d of the centerline is defined as $s_i = \sum_{j=1}^{i} \|p_j - p_{j-1}\|_2$, and the total lane length is $L_l = s_{n_l}$. 
Lane width $w_l$ is estimated as the average lateral distance between paired left and right boundary points, which provides the necessary geometric attributes for discretization.  

The dynamic component is described as a set of traffic participants $\mathcal{A} = \{a_1, a_2, \dots, a_{|\mathcal{A}|}\}$. 
Each agent $a_i$ evolves in discrete time steps $t$, and its instantaneous state is defined as $x_i^t = (p_i^t, v_i^t, \theta_i^t)$, where $p_i^t \in \mathbb{R}^2$ denotes the position, $v_i^t \in \mathbb{R}$ the speed magnitude, and $\theta_i^t \in [-\pi,\pi]$ the heading orientation. 
The trajectory of agent $a_i$ is represented as $\tau_i = \{x_i^0, x_i^1, \dots, x_i^T\}$, where $T$ denotes the %
horizon.

\subsection{Lane Gridification and Occupancy}
To transform the continuous lane geometry into a structured representation, we adopt a lane gridification process. 
From the Argoverse 2~\cite{wilson2023argoverse} HD map, we extract lane centerline, lateral boundaries, connectivity relations, and categorical attributes (e.g. straight, left-turn, right-turn). 
The cumulative arc-length parameterization provides the lane length, while the lateral boundary distances yield an estimate of the lane width.  

For discretizations, each lane is uniformly partitioned along its arc-length. Given a lane length $L_l$, the number of grid cells is defined as:
\[
N_l = \Big\lceil \frac{L_l}{\Delta s} \Big\rceil, \quad \Delta s = 4.0 \,\text{m},
\]
where $\Delta s$ denotes the longitudinal grid resolution. This ensures at least one grid cell per lane and adapts the final cell size when $L_l$ is not divisible by $\Delta s$. The center of the $i$-th grid cell is computed via arc-length parameterization and piecewise linear interpolation, ensuring uniform placement of cell centers along the lane.  
Each cell $g_{l,j}$ is associated with semantic attributes including lane identifier $id_l$, index $j$, lane type $\text{\em type}_l$, and occupancy state $\sigma_{l,j}^t$. 
Connections between cells are introduced to form a grid-level topology, with longitudinal links capturing intra-lane progression, lateral links describing admissible lane changes, %
and inter-lane links %
representing merges or splits.  

The occupancy of each grid cell evolves over time is defined as:
\[
\sigma_{l,j}^t =
\begin{cases}
0, & \text{free cell},\\
1, & \text{occupied by original agents},\\
2, & \text{occupied by generated agents}.
\end{cases}
\]
Their value assignments follow mutual exclusion rules. %
This also implies that generated agents can be placed in any unoccupied grid cell ($\sigma_{l,j}^t = 0$).
The occupancy by the original agents is determined by the ground-truth Argoverse 2 trajectories via hierarchical distance-based mapping, while the occupancy of the generated agents is updated based on %
their simulated paths. 
Conflict resolution prioritizes original agents, while generated agents negotiate conflicts among themselves using behavior-driven arbitration.

\input{sec/algo}

\subsection{Agent Policy}

Unlike the original agents, whose maneuvers are specified by the dataset, the generated agents require the generators to define their action policies for producing specific trajectories. 
On the discrete grid topology $\mathcal{G} = \{g_{l,j}\}$, each generated agent $a_i \in \mathcal{A}$ moves sequentially from one grid cell to another according to its policy. 
The grid cell currently occupied by agent $a_i$ is denoted as $g_i^t = g_{l,j}$, where $l$ is the lane index and $j$ is the longitudinal index along lane $l$.  
The transition of an agent’s status refers to the joint update of both its grid occupancy and its continuous motion state:
\[
(g_i^t, x_i^t) \;\;\longrightarrow\;\; (g_i^{t+1}, x_i^{t+1}),
\]
which is determined by the agent $\text{policy}_i \in \{\text{straight}, \text{left turn}, \text{right turn}, \text{lane change}, \text{overtake}\}$, and the occupancy state $\sigma_{l,j}^t$ of the surrounding grid cells. 
Lane change and overtaking policy are event-driven and are triggered only when specific conditions are met. In particular, we introduce a trigger time $t_{\text{trigger}}^i$ for agent $a_i$, after which the feasibility of switching to an adjacent lane is evaluated. Therefore, the target grid $g_i^{t+1}$ is determined by both the temporal trigger condition $t \ge t_{\text{trigger}}^i$ and the spatial feasibility checks.  
In this way, the discrete grid transition $g_i^t \rightarrow g_i^{t+1}$ and the continuous state update $x_i^t \rightarrow x_i^{t+1}$ are coupled to ensure that each agent follows dynamically feasible and collision-free trajectories.

\subsubsection{\textbf{Lane Change}}
For agent $a_i$, if $t \ge t_{\text{trigger}}^i$, the lane-change feasibility is evaluated. At this moment, the agent first checks whether there exists a valid adjacent lane $l'$ that is parallel and allows lane-changing in the current driving direction. If such a lane exists, the candidate target grid is $g_{l',j}$, which is aligned with the current longitudinal index $j$ of lane $l$.  
Feasibility is further checked by inspecting the local neighborhood of $g_{l',j}$ to ensure that no existing agents occupy the target or adjacent cells, i.e.,
\[
\sigma_{l',j+k}^t = 0, \quad \forall k \in \{-1,0,1\}.
\]
Additionally, the remaining lane $l'$ length must exceed a minimum threshold to guarantee a safe maneuver completion.  

The state transition is then defined as:
\[
g_i^{t+1} =
\begin{cases}
g_{l', j}, & \text{if lane change feasible},\\
g_{l, j+1}, & \text{otherwise keep straight}.
\end{cases}
\]

\subsubsection{\textbf{Overtaking}}
The agent scans its forward observation horizon $d_{\text{obs}}$ as:
\[
\exists g_{l,j+k}, \; k \le \frac{d_{\text{obs}}}{\Delta s}, \quad \sigma_{l,j+k}^t \in \{1,2\} \;\;\Rightarrow\;\; \text{overtake evaluation}.
\]

Feasibility is determined by three conditions (1) the availability of a free overtaking lane $l'$, (2) safe longitudinal gaps:
\[
\Delta d_{\text{ahead}} \ge \Delta d_{\text{safe}}^{\text{front}}, \quad
\Delta d_{\text{behind}} \ge \Delta d_{\text{safe}}^{\text{rear}},
\]
and (3) existence of a continuous overtaking corridor as:
\[
m\Delta s \ge d_{\text{overtake}}.
\]
If all conditions are satisfied, the maneuver proceeds in three stages, (1) the agent enters the overtaking lane at $g_{l',j}$, (2) it continues straight within the overtaking lane at $g_{l',j+k}$, (3) it returns to the original lane at $g_{l,j+k'}$ with $k' > k$.

\subsubsection{\textbf{Straight, Left Turn and Right Turn}} 

On \emph{straight lanes}, agents typically maintain an approximately constant velocity, with minor fluctuations caused by collision avoidance or car-following rules, expressed as 
\(
v(t+\Delta t) \approx v(t)
\), 
where $\Delta t$ denotes the simulation step length.  
On \emph{left-turn lanes}, agents execute progressive deceleration when entering, maintain reduced velocity while turning, and accelerate again after exiting. The velocity profile is modeled as:
\[
v(t) = v_0 \cdot f_{\text{left}}(\alpha(t)),
\]
where $v_0$ is the entry velocity at the lane entrance, $\alpha(t) \in [0,1]$ is the normalized progress along the lane defined by $\alpha(t) = s(t)/L_l$ with $s(t)$ the arc-length position and $L_l$ the total lane length, and $f_{\text{left}}(\alpha)$ is a shaping function that enforces slower speeds in the early and middle segments of the turn.  
On \emph{right-turn lanes}, the maneuver is similar but with milder speed reduction:
\[
v(t) = v_0 \cdot f_{\text{right}}(\alpha(t)), \quad f_{\text{right}}(\alpha) > f_{\text{left}}(\alpha),
\]
where $f_{\text{right}}(\alpha)$ imposes a weaker speed reduction compared to $f_{\text{left}}(\alpha)$, reflecting the smaller curvature and shorter duration of right turns.

\subsection{Trajectory Generation}
Trajectory generation synchronizes temporal discretizations, grid transitions, and conflict management. For agent $a_i$, the dwell time per grid is defined as:
\[
\Delta t_i = \frac{\Delta s}{v_i}, \qquad n_i = \Big\lceil \frac{\Delta t_i}{\delta} \Big\rceil,
\]
where $\Delta s$ is the grid resolution, $v_i$ is the current speed, $\delta$ is the simulation step length, and $n_i$ is the number of steps needed to move one grid.  

The update rule for agent state is expressed as:
\[
x_i^{t+n_i} = f(x_i^t, g_i^{t+1}, \text{policy}_i),
\]
where
\[
\text{policy}_i \in 
\begin{aligned}
\{&\text{straight}, \ \text{left turn}, \ \text{right turn}, \\
  &\text{lane change}, \ \text{overtake}\}.
\end{aligned}
\]
After each transition, the previous grid is released, the new one occupied, and the trajectory $(p_i^t, v_i^t, \theta_i^t)$ logged.  

Conflict resolution includes three cases,   
(1) agent–agent conflicts, where $\sigma=1 \Rightarrow$ reject transition,  
(2) agent–agent direct conflicts, resolved by priority rules (executing $>$ pending, closer $>$ farther),  
(3) future conflicts, where predicted horizon trajectories $\{\hat{x}_i^{t+h}\}_{h=1}^H$ are checked for intersections. If collisions are predicted, agents adjust timing or reroute to avoid unsafe interactions.

\subsection{Trajectory Smoothing}
Finally, discrete grid paths are mapped into continuous, dynamically feasible trajectories in Frenet~\cite{werling2010optimal} coordinates. 
For agent $a_i$, its Cartesian position $p_i^t \in \mathbb{R}^2$ can be expressed in Frenet coordinates $(s_i^t, d_i^t)$, 
where $s_i^t$ is the continuous longitudinal arc-length along the reference line $\mathbf{r}(s)$, 
and $d_i^t$ is the lateral offset. 
The inverse transform is defined as:
\[
p_i^t = \mathbf{r}(s_i^t) + d_i^t \cdot \mathbf{n}(s_i^t),
\]
with $\mathbf{n}(s)$ denoting the unit normal at $\mathbf{r}(s)$.  
In Frenet space, the trajectory of agent $a_i$ is parameterized by cubic polynomials as follows:
\[
s_i(t) = a_0 + a_1 t + a_2 t^2 + a_3 t^3,
\]
\[
d_i(t) = b_0 + b_1 t + b_2 t^2 + b_3 t^3,
\]
where the coefficients $\{a_k, b_k\}$ are determined from boundary conditions including position $p_i^t$, velocity $v_i^t$, and acceleration (derived from velocity differences).  
To ensure dynamic feasibility, we further impose curvature and lateral acceleration constraints. 
The curvature of agent $a_i$’s 2D trajectory $\tau_i(t) = p_i^t = (p_{i,x}(t), p_{i,y}(t))$ is expressed as:
\[
\kappa_i(t) = \frac{|\dot{p}_{i,x}(t)\ddot{p}_{i,y}(t) - \dot{p}_{i,y}(t)\ddot{p}_{i,x}(t)|}{\big(\dot{p}_{i,x}(t)^2+\dot{p}_{i,y}(t)^2\big)^{3/2}},
\]
which must satisfy:
\[
\kappa_i(t) \leq \kappa_{\max} = \frac{1}{R_{\min}},
\]
where $R_{\min}$ is the minimum turning radius.  
The lateral acceleration of agent $a_i$ is defined as:
\[
a_{y}^i(t) = \kappa_i(t) \cdot (v_i^t)^2,
\]
and must remain below $a_{y,\max}$ to ensure safety and comfort, 
where $a_{y,\max}$ is the maximum allowable lateral acceleration.

%% file: sec/algo.tex
\begin{algorithm}[t]
\caption{Agent Decision}
\label{alg:agent_decision}
\textbf{Input:} Grid topology $\mathcal{G} = \{g_{l,j}\}$, agent $a_i$ at time $t$, trigger time $t_{\text{trigger}}^i$, observation horizon $d_{\text{obs}}$, grid resolution $\Delta s$ \\
\textbf{Output:} Next grid state $g_i^{t+1}$ and policy $\text{policy}_i$

\begin{algorithmic}[1]
\State Initialize $\text{policy}_i \gets$ straight;
\State Observe local occupancy $\sigma_{l,j}^t$ and neighboring grids;

\If{$t \geq t_{\text{trigger}}^i$ \&\& adjacent lane $l'$ exists \&\& $g_{l',j}=0$}
    \State $g_i^{t+1} \gets g_{l',j+1}$; 
    \State $\text{policy}_i \gets$ \underline{lane change};
\ElsIf{$\exists g_{l,j+k}, k \leq d_{\text{obs}}/\Delta s$ with $\sigma_{l,j+k}^t \in \{1,2\}$}
    \If{%
        $\Delta d_{\text{ahead}} \geq \Delta d_{\text{safe}}^{\text{front}}$, 
        \&\& $\Delta d_{\text{behind}} \geq \Delta d_{\text{safe}}^{\text{rear}}$, 
        \&\& $m\Delta s \geq d_{\text{overtake}}$}
        \State Execute enter–pass–return sequence in lane $l'$;
        \State $\text{policy}_i \gets$ \underline{overtake};
    \Else
        \State $g_i^{t+1} \gets g_{l,j+1}$; 
    \EndIf
\Else
    \If{$\text{type}_l =$ straight}
        \State Maintain $v(t+\Delta t) \approx v(t)$;
        \State $\text{policy}_i \gets$ \underline{straight};
    \ElsIf{$\text{type}_l =$ left-turn}
        \State Update $v(t) = v_0 \cdot f_{\text{left}}(\alpha(t))$;
        \State $\text{policy}_i \gets$ \underline{left turn};
    \ElsIf{$\text{type}_l =$ right-turn}
        \State Update $v(t) = v_0 \cdot f_{\text{right}}(\alpha(t))$;
        \State $\text{policy}_i \gets$ \underline{right turn};
    \EndIf
\EndIf

\State \textbf{Return} $g_i^{t+1}, \text{policy}_i$
\end{algorithmic}
\end{algorithm}

%% file: sec/4_experiments.tex
\section{Experiments}

\subsection{Experiment Setup}
\textbf{Datasets.}
We evaluate the effectiveness of our proposed HiD$^{2}$ method on Argoverse Motion Forecasting Dataset 1.1~\cite{chang2019argoverse} and Argoverse 2~\cite{wilson2023argoverse}. The Argoverse 1 dataset contains 323,557 real-world driving scenarios, each with 5-second sequences sampled at 10 Hz. The Argoverse 2 dataset contains 250,000 scenarios, each with 11-second sequences sampled at 10 Hz. Using HiD$^{2}$, for Argoverse 2, we generate $55{,}000$ new high-density scenarios and $10{,}000$ additional scenarios that explicitly involve complex driving interactions, including lane changes and overtaking, as well as left and right turns. For Argoverse 1, we generate $20{,}000$ high-density scenarios and $1{,}000$ interaction scenarios.

\begin{table}[t]
    \caption{Comparison of HiD$^{2}$ with the original Argoverse 1 and Argoverse 2 datasets. For all metrics, lower values indicate more realistic and safer trajectories.}
    \centering
    \scalebox{0.98}{
    \setlength{\tabcolsep}{5 pt} %
    \renewcommand{\arraystretch}{1.1} %
    \resizebox{0.48\textwidth}{!}{ 
        \begin{tabular}{l | c c c c c}
            \toprule
             & LO$\downarrow$ 
             & LA$\downarrow$ 
             & JE$\downarrow$ 
             & SCR$\downarrow$ 
             & ORR$\downarrow$  \\
            \midrule
            Argoverse 2~\cite{wilson2023argoverse} &  1.648
                    &  0.476
                    &  8.106
                    &  0.072
                    &  0.142 \\
            HiD$^{2}$ (Ours) & 1.465
                    & 0.486
                    & 7.801
                    & 0.051
                    & 0.114 \\

            \midrule
            Argoverse 1~\cite{chang2019argoverse} &  1.414
                    &  0.838
                    &  11.192
                    & 0.030
                    & 0.009  \\
            HiD$^{2}$ (Ours) & 1.378
                    & 0.976
                    & 11.031
                    & 0.020
                    &  0.007\\
            \bottomrule
        \end{tabular}
        }
    }
    \label{tab:safety_argo2}
\end{table}

\begin{figure}[H]
    \centering
    \includegraphics[width=0.48\textwidth]{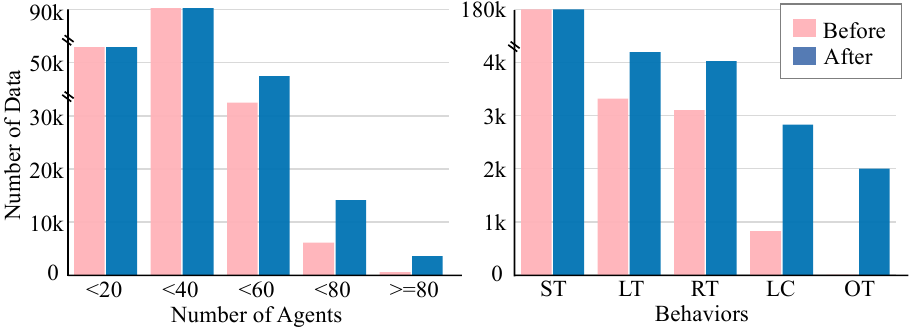}
    \caption{Comparison of dataset distributions before (pink) and after (blue) using HiD$^{2}$ for data generation. \textbf{Left}: distribution of scenarios across different agent density levels (increase in high-density cases). \textbf{Right}: distribution of scenarios with different driving behaviors (enriches the occurrence of complex interactions). }
    
    \label{fig:diversity}
\end{figure}

\subsection{Realism and Safety}

To evaluate the realism and safety of the generated trajectories, we measure both kinematic consistency and scenario-level safety statistics. Specifically, for agent-level realism, we compare several widely used motion quantities~\cite{yang2025trajectory}: 

\begin{table*}[t]
    \caption{Performance comparison of models trained on either Argoverse 2 (random 50k samples) or HiD$^{2}$ and evaluated on both the Argoverse 2 (random 5k samples) and HiD$^{2}$ validation sets. }
    \centering
    \setlength{\tabcolsep}{3 pt}
    \renewcommand{\arraystretch}{1.1}
        \begin{tabular}{l|cc|ccc|ccc}
            \toprule
            \multirow{2}{*}{Method} & \multicolumn{2}{c|}{Dataset} & \multicolumn{3}{c|}{Argoverse 2} & \multicolumn{3}{c}{HiD$^{2}$} \\
            \cmidrule(lr){2-3} \cmidrule(lr){4-6} \cmidrule(lr){7-9}
             & Argoverse 2~\cite{wilson2023argoverse} & HiD$^{2}$ & 
             minADE$\downarrow$& minFDE$\downarrow$& \phantom{x}MR$\downarrow$\phantom{x} &
             minADE$\downarrow$& minFDE$\downarrow$& \phantom{x}MR$\downarrow$\phantom{x} \\
            \midrule
            \multirow{2}{*}{QCNet~\cite{zhou2023query}} & \checkmark &  & 0.845 & 1.573 & 0.228 & 0.925 & 1.701 & 0.254 \\
            &  & \checkmark & 0.849 & 1.584 & 0.223 & 0.835 & 1.537 & 0.211 \\
            \midrule
            \multirow{2}{*}{DeMo~\cite{zhang2024decoupling}} &  
            \checkmark &  & 0.745 & 1.511 & 0.208 & 0.823 & 1.645 & 0.234 \\
            &  & \checkmark & 0.765 & 1.548 & 0.221 & 0.779 & 1.530 & 0.212 \\
            \bottomrule
        \end{tabular}

    \label{tab:argo2_and_HiD$^{2}$}
\end{table*}

\begin{table*}[t]
    \caption{Performance comparison of models trained on Argoverse 2 alone and on the combined Argoverse 2 and HiD$^{2}$ dataset, evaluated under increasingly dense scenarios with more than 50, 70, and 90 interacting agents. }
    \centering
    \setlength{\tabcolsep}{3 pt}
    \renewcommand{\arraystretch}{1.1}
        \begin{tabular}{l|cc|ccc|ccc|ccc}
            \toprule
            \multirow{2}{*}{Method} & \multicolumn{2}{c|}{Train Dataset} 
            & \multicolumn{3}{c|}{Agent$>$50} 
            & \multicolumn{3}{c|}{Agent$>$70} 
            & \multicolumn{3}{c}{Agent$>$90} \\
            \cmidrule(lr){2-3} 
            \cmidrule(lr){4-6} \cmidrule(lr){7-9} \cmidrule(lr){10-12}
             & Argoverse 2~\cite{wilson2023argoverse} & HiD$^{2}$ 
             & minADE$\downarrow$ & minFDE$\downarrow$ & \phantom{x}MR$\downarrow$\phantom{x} 
             & minADE$\downarrow$ & minFDE$\downarrow$ & \phantom{x}MR$\downarrow$\phantom{x}
             & minADE$\downarrow$ & minFDE$\downarrow$ & \phantom{x}MR$\downarrow$\phantom{x} \\
            \midrule
            \multirow{2}{*}{QCNet~\cite{zhou2023query}} 
            & \checkmark &  & 0.728 & 1.237 & 0.160 & 0.734 & 1.257 & 0.166 & 0.745 & 1.243 & 0.169 \\
            & \checkmark & \checkmark & \textbf{0.723} & \textbf{1.214} & \textbf{0.152} & \textbf{0.730} & \textbf{1.206} & \textbf{0.160} & \textbf{0.731} & \textbf{1.217} & \textbf{0.161} \\
            \midrule
            \multirow{2}{*}{DeMo~\cite{zhang2024decoupling}} 
            & \checkmark &  & 0.674 & 1.320 & 0.167 & 0.688 & 1.327 & 0.173 & 0.696 & 1.357 & 0.176 \\
            & \checkmark & \checkmark & \textbf{0.663} & \textbf{1.291} & \textbf{0.164} & \textbf{0.678} & \textbf{1.293} & \textbf{0.167} & \textbf{0.641} & \textbf{1.219} & \textbf{0.153} \\
            \bottomrule
        \end{tabular}
    \label{tab:more_agent}
\end{table*}

\textbf{(i) Longitudinal acceleration (LO)}, defined as:
\[
LO = |a_{\parallel}| = |a_x \cos\theta + a_y \sin\theta|,
\]
where $(a_x,a_y)$ is the acceleration vector and $\theta$ is the vehicle heading.  

\textbf{(ii) Lateral acceleration (LA)}, defined as:
\[
LA = |a_{\perp}| = |-a_x \sin\theta + a_y \cos\theta|.
\]

\textbf{(iii) Jerk (JE)}, defined as the magnitude of the time derivative of the acceleration vector:
\[
JE = \sqrt{\left(\tfrac{da_x}{dt}\right)^2 + \left(\tfrac{da_y}{dt}\right)^2}.
\]
These metrics characterize the smoothness and naturalness of individual vehicle motions.  

In addition to agent-level realism, we further consider scene-level safety~\cite{lin2025causal} indicators:  

\textbf{(i) Scenario collision rate (SCR)}, defined as:
\[
SCR = \frac{1}{|\mathcal{S}|} \sum_{s \in \mathcal{S}} 
      \frac{\#\{\text{colliding vehicles in } s\}}
           {\#\{\text{total vehicles in } s\}},
\]
where two vehicles are considered colliding if their oriented bounding boxes overlap above a small IoU threshold.  

\textbf{(ii) Off-road rate (ORR)}, defined as:
\[
ORR = \frac{\#\{p_i \notin \mathcal{D}\}}{\#\{p_i\}},
\]
where $\mathcal{D}$ denotes the drivable area polygons from the HD map.  
These indicators capture whether generated trajectories remain physically plausible and safe at the scene level.

As shown in Table ~\ref{tab:safety_argo2}, the data generated using HiD$^{2}$ maintains high statistical consistency with the original Argoverse 2 dataset. This demonstrates that the generated trajectories simulate smoother and more natural single-vehicle dynamics at the agent level, while maintaining low collision and off-road violation rates at the scene level. Notably, the inclusion of more interacting vehicles in the generated scenarios broadens the data distribution, effectively supplementing the scarce long-tail, high-density scenario in the original dataset.

\begin{figure*}[t]
    \centering
    \includegraphics[width=\textwidth]{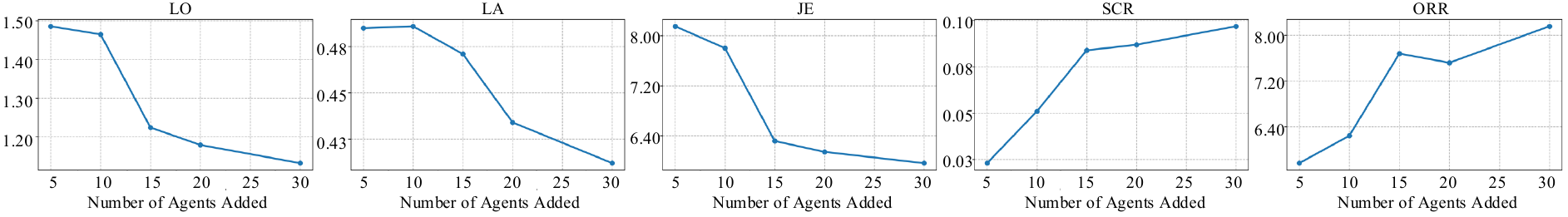}
    \caption{Effect of adding different numbers of agents into the scenarios. From left to right are LO, LA, JE, SCR, and ORR.} 
    \label{fig:add_diff_agent}
\end{figure*}

In addition, we further investigate the effect of adding different numbers of agents into the scenarios, as shown in Table~\ref{fig:add_diff_agent}. The results show that when only a small number of agents are added, the LO, LA, and JE values remain higher due to the relatively low scenario density. At the same time, the SCR and ORR values are lower, indicating safer conditions. As more agents are introduced, the trajectories remain stable and smooth with decreasing LO, LA, and JE, while SCR and ORR increase moderately. This reflects the fact that high-density scenarios naturally carry more collision and off-road risks; nevertheless, our method is able to keep these rates at a relatively low level, demonstrating its robustness in generating realistic yet safe congested scenarios.

\subsection{Diversity}
To enhance the diversity of training data, we leverage original high-density scenarios with more than 40 agents as templates and augment them by adding 10 additional agents into each scenario, thereby increasing the proportion of scenarios with more than 50 agents. 
In addition to increasing scenario density, we also generate new scenarios that contain underrepresented driving behaviors. Specifically, ST (straight) corresponds to simple driving, while LT (left turn), RT (right turn), LC (lane change), and OT (overtaking) capture more diverse and interactive maneuvers. Our generation process not only enrich the lane-change and overtaking behaviors but also increases the left-turn and right-turn maneuvers, which are relatively scarce in the original datasets.
Figure~\ref{fig:diversity} illustrates the comparison before and after augmentation. On the left, the distribution shift shows a clear increase in the number of scenarios with more than 40 agents. On the right, the behavioral distribution indicates noticeable gains in turning and complex interaction scenarios, while preserving the balance of simpler straight-driving behaviors.

\subsection{Trajectory Prediction Performance}
\textbf{Metrics.}
We evaluate prediction performance with the standard metrics: minimum Average Displacement Error (minADE), minimum Final Displacement Error (minFDE), and Miss Rate (MR). Models can output up to 6 trajectories per agent. minADE measures the average distance between the best prediction and the ground truth over all future steps, minFDE measures the distance at the final step, and MR is the fraction of cases where the closest predicted endpoint is more than 2.0 meters from the ground-truth endpoint.

\textbf{Separate Evaluation on Argoverse 2 and HiD$^{2}$.} To further validate the effectiveness of HiD$^{2}$, we conduct trajectory prediction experiments using QCNet~\cite{zhou2023query} and DeMo~\cite{zhang2024decoupling}. Specifically, we generate $55{,}000$ HiD$^{2}$ scenarios and split them into a non-overlapping training set of $50{,}000$ samples and a validation set of $5{,}000$ samples. For fair comparison, we also randomly sample $50{,}000$ scenarios from the original Argoverse 2 dataset and use them to train QCNet and DeMo as baselines. The models are then evaluated on both the $5{,}000$ validation samples drawn from Argoverse 2 and the $5{,}000$ HiD$^{2}$ validation samples.  

As shown in Table~\ref{tab:argo2_and_HiD$^{2}$}, models trained on HiD$^{2}$ achieve slightly worse performance on the Argoverse 2 validation set compared to those trained directly on Argoverse 2. 
This slight degradation is expected, since HiD$^{2}$ emphasizes high-density scenarios with more than 50 interacting agents, whereas the majority of Argoverse 2 scenarios are low-density. The distributional gap makes HiD$^{2}$-trained models less specialized for sparse-traffic cases. Nevertheless, when evaluated on HiD$^{2}$’s own validation set, both QCNet and DeMo trained on HiD$^{2}$ consistently outperform their Argoverse 2 counterparts. Specifically, for QCNet, minADE decreases by -9.7\%, minFDE by -9.6\%, and MR by -16.9\%. For DeMo, the improvements are similar, with minADE reduced by -5.3\%, minFDE by -7.0\%, and MR by -9.4\%. These results show HiD$^{2}$ enriches the dataset with underrepresented complex and safety-critical behaviors.

\textbf{Combined Evaluation on Argoverse 2 and HiD$^{2}$.} 
To further examine the benefits of combining HiD$^{2}$ with the original dataset, we evaluate models trained on Argoverse 2 only and Argoverse 2 + HiD$^{2}$ under dense scenarios with more than 50, 70, and 90 agents, as shown in Table~\ref{tab:more_agent}. When trained on Argoverse 2, both QCNet and DeMo exhibit clear performance degradation as the number of interacting agents grows. In contrast, incorporating HiD$^{2}$ consistently improves generalization in these challenging settings, yielding lower minADE, minFDE and MR across all density thresholds. Specifically, for agents more than 50, QCNet achieves reductions of 1.9\% in minFDE, and 5.0\% in MR, while DeMo reduces minFDE by 2.2\%, and MR by 1.8\%. For agents more than 70, QCNet achieves 4.1\%, and 3.6\% reductions in minFDE, and MR, respectively, whereas DeMo achieves decreases of 2.6\%, and 3.5\%. For agents more than 90, QCNet lowers minFDE by 2.1\%, and MR by 4.7\%, while DeMo lowers them by 10.2\%, and 13.1\%. 
The results demonstrate that HiD$^{2}$ provides complementary high-density scenarios that substantially enhance model robustness under high-density scenarios.

\begin{table}[t]
    \caption{Ablation comparison using different training sources. HiD$^{2}$ consistently improves both accuracy (LO, LA, JE) and safety (SCR, ORR) over Argoverse 2.}
    \centering
    \setlength{\tabcolsep}{5 pt}
    \renewcommand{\arraystretch}{1.2}
    \scalebox{0.99}{
    \setlength{\tabcolsep}{2 pt} %
    \renewcommand{\arraystretch}{1.1} %
    \resizebox{0.48\textwidth}{!}{ 
        \begin{tabular}{c c c | c c c c c}
            \toprule
            Topology & Collision & Smooth & LO$\downarrow$ & LA$\downarrow$ & JE$\downarrow$ & SCR$\downarrow$ & ORR$\downarrow$ \\
            \midrule
            \ding{55} &            &            & 1.491 & 0.483 & 7.842  & 0.101  & 0.225 \\
                       & \ding{55} &            & 1.449 & 0.433 & 7.087 & 0.282 & 0.125 \\
                       &           & \ding{55}  & 2.379 & 1.188 & 9.156 & 0.072 & 0.141 \\
            \bottomrule
        \end{tabular}
        }
    }

    \label{tab:ablation}
\end{table}

\subsection{Ablation Study}

We conduct ablation studies for the multi-agent trajectory generation task on the Argoverse 2 dataset to systematically assess the necessity of each component in our framework, with results summarized in Table~\ref{tab:ablation}. The ablation results reveal clear evidence that each module provides complementary benefits, and removing any one of them leads to noticeable degradation in either safety or motion quality.

When lane topology analysis is removed, the system loses structural guidance from lane connectivity, which is essential for ensuring that agents follow realistic driving flows. Without this guidance, generated agents tend to drift into incorrect lanes or even leave the drivable area entirely. This structural deficiency directly translates into safety issues: SCR rises by +98\% from 0.051 to 0.101 and ORR by +97\% from 0.114 to 0.225, showing that agents are more likely to collide or deviate off-road. Disabling collision detection leads to the most severe safety degradation. Without explicit collision checking, the system fails to filter out unsafe trajectories that intersect with existing agents. SCR surges by +453\% from 0.051 to 0.282, while ORR still increases by +10\% from 0.114 to 0.125. Besides, removing trajectory smoothing strongly impacts motion quality. Without Frenet-based refinement, the generated trajectories are piecewise linear at the grid level, lacking continuous curvature. As a result,  LO increases by +62\% from 1.465 to 2.379, LA by +139\% from 0.496 to 1.188, and JE by +17\% from 7.801  9.156, leading to jerky and unnatural motion. In contrast, SCR and ORR remain relatively stable.

%% file: sec/5_conclusion.tex
\section{Conclusion}

This paper tackled the long-tail problem in trajectory prediction, characterized by imbalanced agent density and limited interaction diversity. We proposed HiD$^{2}$ a unified trajectory generation framework that discretizes real maps into structured grids, incorporates behavior-aware planning, and applies Frenet-based smoothing with dynamic feasibility constraints. HiD$^{2}$ is the first to generate high-density, multi-interaction scenarios directly on real maps without relying on existing data distributions, producing realistic and safe trajectories with rare behaviors such as lane changes, overtaking, left turn and right turn. Experiments on Argoverse 1 and Argoverse 2 demonstrate that the generated data not only enriches critical long-tail scenarios but also, when combined with real data, substantially improves the generalization of existing prediction models in high-density environments.

%% file: root.bbl
\begin{thebibliography}{10}
\providecommand{\url}[1]{#1}
\csname url@samestyle\endcsname
\providecommand{\newblock}{\relax}
\providecommand{\bibinfo}[2]{#2}
\providecommand{\BIBentrySTDinterwordspacing}{\spaceskip=0pt\relax}
\providecommand{\BIBentryALTinterwordstretchfactor}{4}
\providecommand{\BIBentryALTinterwordspacing}{\spaceskip=\fontdimen2\font plus
\BIBentryALTinterwordstretchfactor\fontdimen3\font minus \fontdimen4\font\relax}
\providecommand{\BIBforeignlanguage}[2]{{%
\expandafter\ifx\csname l@#1\endcsname\relax
\typeout{** WARNING: IEEEtran.bst: No hyphenation pattern has been}%
\typeout{** loaded for the language `#1'. Using the pattern for}%
\typeout{** the default language instead.}%
\else
\language=\csname l@#1\endcsname
\fi
#2}}
\providecommand{\BIBdecl}{\relax}
\BIBdecl

\bibitem{wang2025risk}
Q.~Wang, D.~Xu, G.~Kuang, C.~Lv, S.~E. Li, and B.~Nie, ``Risk-aware vehicle trajectory prediction under safety-critical scenarios,'' \emph{IEEE Transactions on Intelligent Transportation Systems}, 2025.

\bibitem{yang2024towards}
K.~Yang, S.~Li, Y.~Chen, D.~Cao, and X.~Tang, ``Towards safe decision-making for autonomous vehicles at unsignalized intersections,'' \emph{IEEE Transactions on Vehicular Technology}, 2024.

\bibitem{vaswani2017attention}
A.~Vaswani, N.~Shazeer \emph{et~al.}, ``Attention is all you need,'' \emph{Advances in neural information processing systems}, vol.~30, 2017.

\bibitem{ngiam2021scene}
J.~Ngiam, B.~Caine \emph{et~al.}, ``Scene transformer: A unified architecture for predicting multiple agent trajectories,'' \emph{arXiv preprint arXiv:2106.08417}, 2021.

\bibitem{wilson2023argoverse}
B.~Wilson, W.~Qi, T.~Agarwal, J.~Lambert, J.~Singh, S.~Khandelwal, B.~Pan, R.~Kumar, A.~Hartnett, J.~K. Pontes \emph{et~al.}, ``Argoverse 2: Next generation datasets for self-driving perception and forecasting,'' \emph{arXiv preprint arXiv:2301.00493}, 2023.

\bibitem{ettinger2021large}
S.~Ettinger, S.~Cheng \emph{et~al.}, ``Large scale interactive motion forecasting for autonomous driving: The waymo open motion dataset,'' in \emph{Proceedings of the IEEE/CVF International Conference on Computer Vision}, 2021, pp. 9710--9719.

\bibitem{chang2019argoverse}
M.-F. Chang, J.~Lambert, P.~Sangkloy, J.~Singh, S.~Bak, A.~Hartnett, P.~C. De~Wang, S.~Lucey, D.~Ramanan, and J.~Hays, ``Argoverse: 3d tracking and forecasting with rich maps,'' in \emph{Proceedings of the IEEE/CVF conference on computer vision and pattern recognition}, 2019, pp. 8748--8757.

\bibitem{huang2018apolloscape}
X.~Huang \emph{et~al.}, ``The apolloscape dataset for autonomous driving,'' in \emph{Proceedings of the IEEE conference on computer vision and pattern recognition workshops}.\hskip 1em plus 0.5em minus 0.4em\relax nill: nill, 2018, pp. 954--960.

\bibitem{caesar2020nuscenes}
H.~Caesar, V.~Bankiti \emph{et~al.}, ``nuscenes: A multimodal dataset for autonomous driving,'' in \emph{Proceedings of the IEEE/CVF conference on computer vision and pattern recognition}, 2020, pp. 11\,621--11\,631.

\bibitem{yang2024sstp}
R.~Yang, Y.~Xu, Y.~Fu, and L.~Su, ``Sstp: Efficient sample selection for trajectory prediction,'' \emph{arXiv preprint arXiv:2409.17385}, 2024.

\bibitem{zhou2022hivt}
Z.~Zhou, L.~Ye, J.~Wang, K.~Wu, and K.~Lu, ``Hivt: Hierarchical vector transformer for multi-agent motion prediction,'' in \emph{Proceedings of the IEEE/CVF Conference on Computer Vision and Pattern Recognition}, 2022, pp. 8823--8833.

\bibitem{tang2024hpnet}
X.~Tang, M.~Kan, S.~Shan, Z.~Ji, J.~Bai, and X.~Chen, ``Hpnet: Dynamic trajectory forecasting with historical prediction attention,'' in \emph{Proceedings of the IEEE/CVF Conference on Computer Vision and Pattern Recognition}, 2024, pp. 15\,261--15\,270.

\bibitem{wang2024towards}
J.~Wang, L.~Su, S.~Han, D.~Song, and F.~Miao, ``Towards safe autonomy in hybrid traffic: Detecting unpredictable abnormal behaviors of human drivers via information sharing,'' \emph{ACM Transactions on Cyber-Physical Systems}, vol.~8, no.~2, pp. 1--25, 2024.

\bibitem{shin2023fill}
J.~Shin, M.~Kang, and J.~Park, ``Fill-up: Balancing long-tailed data with generative models,'' \emph{arXiv preprint arXiv:2306.07200}, 2023.

\bibitem{bansal2023leaving}
H.~Bansal and A.~Grover, ``Leaving reality to imagination: Robust classification via generated datasets,'' \emph{arXiv preprint arXiv:2302.02503}, 2023.

\bibitem{dosovitskiy2017carla}
A.~Dosovitskiy, G.~Ros, F.~Codevilla, A.~Lopez, and V.~Koltun, ``Carla: An open urban driving simulator,'' in \emph{Conference on robot learning}.\hskip 1em plus 0.5em minus 0.4em\relax PMLR, 2017, pp. 1--16.

\bibitem{yang2025trajectory}
K.~Yang, Z.~Guo \emph{et~al.}, ``Trajectory-llm: A language-based data generator for trajectory prediction in autonomous driving,'' in \emph{The Thirteenth International Conference on Learning Representations}, 2025.

\bibitem{feng2022trafficgen}
L.~Feng, Q.~Li, Z.~Peng, S.~Tan, and B.~Zhou, ``Trafficgen: Learning to generate diverse and realistic traffic scenarios,'' \emph{arXiv preprint arXiv:2210.06609}, 2022.

\bibitem{tan2023language}
S.~Tan, B.~Ivanovic, X.~Weng, M.~Pavone, and P.~Kraehenbuehl, ``Language conditioned traffic generation,'' \emph{arXiv preprint arXiv:2307.07947}, 2023.

\bibitem{xia2024language}
J.~Xia, C.~Xu, Q.~Xu, Y.~Wang, and S.~Chen, ``Language-driven interactive traffic trajectory generation,'' \emph{Advances in Neural Information Processing Systems}, vol.~37, pp. 77\,831--77\,859, 2024.

\bibitem{giuliari2021transformer}
F.~Giuliari, I.~Hasan, M.~Cristani, and F.~Galasso, ``Transformer networks for trajectory forecasting,'' in \emph{2020 25th international conference on pattern recognition (ICPR)}.\hskip 1em plus 0.5em minus 0.4em\relax IEEE, 2021, pp. 10\,335--10\,342.

\bibitem{xu2020spatial}
M.~Xu, W.~Dai, C.~Liu, X.~Gao, W.~Lin, G.-J. Qi, and H.~Xiong, ``Spatial-temporal transformer networks for traffic flow forecasting,'' \emph{arXiv preprint arXiv:2001.02908}, 2020.

\bibitem{zhou2023query}
Z.~Zhou, J.~Wang, Y.-H. Li, and Y.-K. Huang, ``Query-centric trajectory prediction,'' in \emph{Proceedings of the IEEE/CVF conference on computer vision and pattern recognition}, 2023, pp. 17\,863--17\,873.

\bibitem{zhang2025decoupling}
B.~Zhang, N.~Song, and L.~Zhang, ``Decoupling motion forecasting into directional intentions and dynamic states,'' \emph{Advances in Neural Information Processing Systems}, vol.~37, pp. 106\,582--106\,606, 2025.

\bibitem{liang2020garden}
J.~Liang, L.~Jiang, K.~Murphy, T.~Yu, and A.~Hauptmann, ``The garden of forking paths: Towards multi-future trajectory prediction,'' in \emph{Proceedings of the IEEE/CVF conference on computer vision and pattern recognition}, 2020, pp. 10\,508--10\,518.

\bibitem{makansi2021exposing}
O.~Makansi, {\"O}.~{\c{C}}i{\c{c}}ek, Y.~Marrakchi, and T.~Brox, ``On exposing the challenging long tail in future prediction of traffic actors,'' in \emph{Proceedings of the IEEE/CVF International Conference on Computer Vision}, 2021, pp. 13\,147--13\,157.

\bibitem{pourkeshavarz2023learn}
M.~Pourkeshavarz, C.~Chen, and A.~Rasouli, ``Learn tarot with mentor: A meta-learned self-supervised approach for trajectory prediction,'' in \emph{Proceedings of the IEEE/CVF International Conference on Computer Vision}, 2023, pp. 8384--8393.

\bibitem{guo2024building}
T.~Guo, T.~Banerjee, R.~Liu, and L.~Su, ``Building real-time awareness of out-of-distribution in trajectory prediction for autonomous vehicles,'' \emph{arXiv preprint arXiv:2409.17277}, 2024.

\bibitem{wang2023fend}
Y.~Wang, P.~Zhang, L.~Bai, and J.~Xue, ``Fend: A future enhanced distribution-aware contrastive learning framework for long-tail trajectory prediction,'' in \emph{Proceedings of the IEEE/CVF conference on computer vision and pattern recognition}, 2023, pp. 1400--1409.

\bibitem{zhou2022long}
W.~Zhou, Z.~Cao \emph{et~al.}, ``Long-tail prediction uncertainty aware trajectory planning for self-driving vehicles,'' in \emph{2022 IEEE 25th International Conference on Intelligent Transportation Systems (ITSC)}.\hskip 1em plus 0.5em minus 0.4em\relax IEEE, 2022, pp. 1275--1282.

\bibitem{maroto2006real}
J.~Maroto, E.~Delso, J.~Felez, and J.~M. Cabanellas, ``Real-time traffic simulation with a microscopic model,'' \emph{IEEE Transactions on Intelligent Transportation Systems}, vol.~7, no.~4, pp. 513--527, 2006.

\bibitem{casas2010traffic}
J.~Casas, J.~L. Ferrer, D.~Garcia, J.~Perarnau, and A.~Torday, ``Traffic simulation with aimsun,'' in \emph{Fundamentals of traffic simulation}.\hskip 1em plus 0.5em minus 0.4em\relax Springer, 2010, pp. 173--232.

\bibitem{papaleondiou2009trafficmodeler}
L.~G. Papaleondiou and M.~D. Dikaiakos, ``Trafficmodeler: A graphical tool for programming microscopic traffic simulators through high-level abstractions,'' in \emph{VTC Spring 2009-IEEE 69th Vehicular Technology Conference}.\hskip 1em plus 0.5em minus 0.4em\relax IEEE, 2009, pp. 1--5.

\bibitem{erdmann2015sumo}
J.~Erdmann, ``Sumo’s lane-changing model,'' in \emph{Modeling Mobility with Open Data: 2nd SUMO Conference 2014 Berlin, Germany, May 15-16, 2014}.\hskip 1em plus 0.5em minus 0.4em\relax Springer, 2015, pp. 105--123.

\bibitem{fellendorf2010microscopic}
M.~Fellendorf and P.~Vortisch, ``Microscopic traffic flow simulator vissim,'' in \emph{Fundamentals of traffic simulation}.\hskip 1em plus 0.5em minus 0.4em\relax Springer, 2010, pp. 63--93.

\bibitem{lopez2018microscopic}
P.~A. Lopez, M.~Behrisch \emph{et~al.}, ``Microscopic traffic simulation using sumo,'' in \emph{2018 21st international conference on intelligent transportation systems (ITSC)}.\hskip 1em plus 0.5em minus 0.4em\relax Ieee, 2018, pp. 2575--2582.

\bibitem{suo2021trafficsim}
S.~Suo, S.~Regalado, S.~Casas, and R.~Urtasun, ``Trafficsim: Learning to simulate realistic multi-agent behaviors,'' in \emph{Proceedings of the IEEE/CVF Conference on Computer Vision and Pattern Recognition}, 2021, pp. 10\,400--10\,409.

\bibitem{tan2021scenegen}
S.~Tan, K.~Wong, S.~Wang, S.~Manivasagam, M.~Ren, and R.~Urtasun, ``Scenegen: Learning to generate realistic traffic scenes,'' in \emph{Proceedings of the IEEE/CVF Conference on Computer Vision and Pattern Recognition}, 2021, pp. 892--901.

\bibitem{xu2022bits}
D.~Xu, Y.~Chen, B.~Ivanovic, and M.~Pavone, ``Bits: Bi-level imitation for traffic simulation,'' \emph{arXiv preprint arXiv:2208.12403}, 2022.

\bibitem{sun2022intersim}
Q.~Sun, X.~Huang, B.~C. Williams, and H.~Zhao, ``Intersim: Interactive traffic simulation via explicit relation modeling,'' in \emph{2022 IEEE/RSJ International Conference on Intelligent Robots and Systems (IROS)}.\hskip 1em plus 0.5em minus 0.4em\relax IEEE, 2022, pp. 11\,416--11\,423.

\bibitem{zhong2022guided}
Z.~Zhong, D.~Rempe \emph{et~al.}, ``Guided conditional diffusion for controllable traffic simulation,'' \emph{arXiv preprint arXiv:2210.17366}, 2022.

\bibitem{zhong2023language}
Z.~Zhong, D.~Rempe, Y.~Chen, B.~Ivanovic, Y.~Cao, D.~Xu, M.~Pavone, and B.~Ray, ``Language-guided traffic simulation via scene-level diffusion,'' in \emph{Conference on robot learning}.\hskip 1em plus 0.5em minus 0.4em\relax PMLR, 2023, pp. 144--177.

\bibitem{werling2010optimal}
M.~Werling, J.~Ziegler, S.~Kammel, and S.~Thrun, ``Optimal trajectory generation for dynamic street scenarios in a frenet frame,'' in \emph{2010 IEEE international conference on robotics and automation}.\hskip 1em plus 0.5em minus 0.4em\relax IEEE, 2010, pp. 987--993.

\bibitem{zhang2024decoupling}
B.~Zhang, N.~Song, and L.~Zhang, ``Decoupling motion forecasting into directional intentions and dynamic states,'' \emph{Advances in Neural Information Processing Systems}, vol.~37, pp. 106\,582--106\,606, 2024.

\bibitem{lin2025causal}
H.~Lin, X.~Huang \emph{et~al.}, ``Causal composition diffusion model for closed-loop traffic generation,'' in \emph{Proceedings of the Computer Vision and Pattern Recognition Conference}, 2025, pp. 27\,542--27\,552.

\end{thebibliography}
